\def\year{2022}\relax
\documentclass[letterpaper]{article} 
\usepackage{aaai22}  
\usepackage{times}  
\usepackage{helvet}  
\usepackage{courier}  
\usepackage[hyphens]{url}  
\usepackage{graphicx} 
\usepackage{amsmath}
\usepackage{natbib}  
\usepackage{caption} 
\DeclareCaptionStyle{ruled}{labelfont=normalfont,labelsep=colon,strut=off} 
\usepackage{algorithm}
\usepackage{algorithmic}

\usepackage{newfloat}
\usepackage{listings}
\floatstyle{ruled}
\newfloat{listing}{tb}{lst}{}
\floatname{listing}{Listing}
\title{Maximize the Exploration of Congeneric Semantics for Weakly Supervised Semantic Segmentation}
\author{
    
Ke Zhang$^{1,3}$, Sihong Chen$^{2}$, Qi Ju$^{2}$, Yong Jiang$^{3}$, Yucong Li$^{2}$,  Xin He$^{1,3}$\\

$^{1}$Tsinghua Berkeley Shenzhen Institute \qquad$^{2}$Tencent \qquad$^{3}$Tsinghua University
}
\affiliations{

}

\usepackage{bibentry}

\begin{document}

\maketitle 

\begin{abstract}

With the increase in the number of image data and the lack of corresponding labels, weakly supervised learning has drawn a lot of attention recently in computer vision tasks, especially in the fine-grained semantic segmentation problem. To alleviate human efforts from expensive pixel-by-pixel annotations, our method focuses on weakly supervised semantic segmentation (WSSS) with image-level tags, which are much easier to obtain. As a huge gap exists between pixel-level segmentation and image-level labels, how to reflect the image-level semantic information on each pixel is an important question. To explore the congeneric semantic regions from the same class to the maximum, we construct the patch-level graph neural network (P-GNN) based on the self-detected patches from different images that contain the same class labels. Patches can frame the objects as much as possible and include as little background as possible. 
The graph network that is established with patches as the nodes can maximize the mutual learning of similar objects. We regard the embedding vectors of patches as nodes, and use transformer-based complementary learning module to construct weighted edges according to the embedding similarity between different nodes. Moreover, to better supplement semantic information, we propose soft-complementary loss functions matched with the whole network structure. We conduct experiments on the popular PASCAL VOC 2012 benchmarks, and our model yields state-of-the-art performance.

\end{abstract}

\section{Introduction}
\noindent With the increase in data volume and the lack of annotations, weakly supervised learning is getting more and more attention as people try to use cheaper labels to deal with complex problems in many application scenarios, like object detection, semantic segmentation in images and videos. Weakly supervised semantic segmentation aims to finish pixel-level semantic segmentation with weak labels, such as bounding boxes, scribbles, and image-level class labels. In this paper, we mainly focus on semantic segmentation with class labels (SSCL) as class labels are the easiest to obtain. The key challenge for SSCL is how to establish connections between category semantics and spatial pixel-level features as it is difficult to generate accurate segmentation results without any spatial information provided.

\begin{figure}[t]
\centering
\includegraphics[width=0.8\linewidth,height=2.7cm]{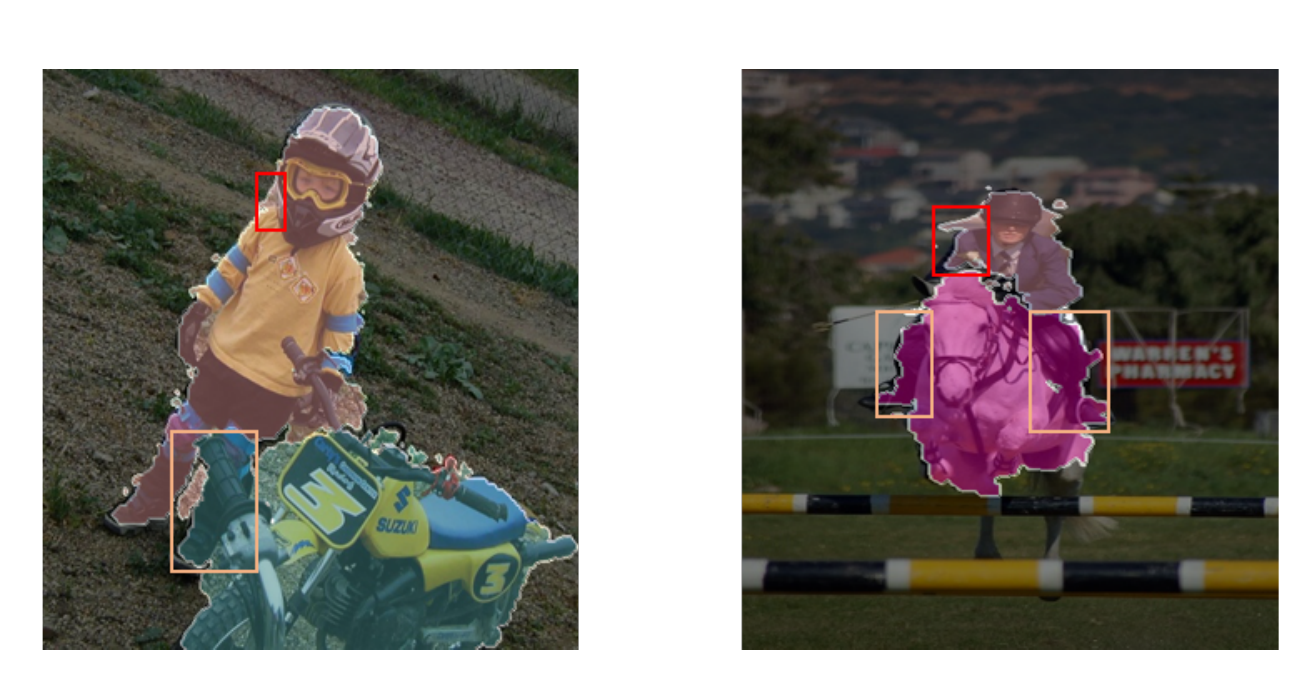} 
\caption{False-positive (the area inside the red boxes) and false-negative samples (the area inside the orange boxes) for the class 'people' in weakly supervised semantic segmentation results.}
\end{figure}

Previous works for SSCL can be classified into two mainstream branches: complete semantic information exploration \cite{sun2020mining,li2020group} and boundary refinement \cite{chen2020weakly}. In this paper, our contributions are concentrated in maximizing the exploration of complete semantic regions. Some boundary refinement methods could also be adopted into our model. In SSCL learning, class activation maps (CAMs) \cite{zhou2016learning} firstly localize objects based on image-level labels. However, the localization maps cannot contain entire objects, like only localizing the head of one person. Some methods use dilated convolutions \cite{wei2018revisiting} or adversarial erasing \cite{zhang2018adversarial} to supplement semantic information based on CAMs in one image. To further explore congeneric semantic relations between different images, \cite{sun2020mining,li2020group} add the attention mechanism between two or more different images to the model. However, their explorations of semantic areas stay at image-level so the segmentation maps are not so accurate and 
the single attention mechanism connection is not stable enough. The main problem of these methods is to mix different information of object foreground and background for training. In reality, the features of the foreground and background and of different objects in the foreground even in the same image can be totally different. Therefore, how to enhance the correlation between congeneric semantics in a more fine-grained manner need to be explored.

To explore the correct semantics in more detail, we trace back to the evaluation metric of semantic segmentation mIoU. In Figure 1, we show examples of false-negative pixels and false-positive pixels. We can find the main reason for generating false-negative samples is the deviation of the model in learning the complete semantic information. For example, when one person stands closer to one motorcycle in Figure 1, the deep learning model judges that the person's one leg is part of the motorcycle because the color of the person's pants is similar to the motorcycle. Meanwhile, false-positive samples are formed due to imprecise edge segmentation as shown in red boxes in Figure 1. Due to false-negative samples occupy the majority of incorrectly classified pixels in SSCL, in this paper, we mainly improve semantic segmentation results by reducing false negative pixels. False-positive samples can be reduced by adding boundary refinement methods into our model. Previous methods establish the attention mechanism for the whole feature maps and ignore the differences in semantic features of different objects in one image, and thus the generated maps always segment foreground pixels into the background if their semantic features are similar, like the pants and the motorbike in the similar color in Figure 1. To solve this problem, we generate a patch-level transformer-based graph neural network to build connections among different patches of different images in the same category. We use patches to frame single objects and establish a graph neural network to enhance mutual semantic learning.

As shown in Figure 2, the input images are from the same class as a mini-batch. We crop every image into different patches and treat each patch as one node in the graph. Taking patches as nodes can carefully separate objects of different categories. Graph neural network based on different patches enhances the mutual semantic learning of the same category. Transformer-based complementary learning module serves to build attention weights for each edge in the graph. In order to build robust connections for different nodes, we adopt a transformer-based structure and make connections through queries, keys, and values. The class embedding vectors are the unique design to make up for the lack of pixel-level label information under weak supervision. For each cropped patch, we add its detected class embedding labels generated by the weakly supervised object detection method \cite{dong2021boosting}, which serves to enhance the semantic information at pixel level. To further match up with the patch-level graph neural network model, we propose soft-complementary loss to complement semantic areas in attention maps generated by the model. Through the combination of these innovations, our model has a good performance on semantic segmentation on PASCAL VOC 2012 dataset.

In summary, our main contributions are three folds:
\begin{itemize}
\item We propose that the main problem affecting weakly supervised semantic segmentation accuracy is the existence of a large number of false-negative samples, which are caused by the weakness of the WSSS model in learning single category semantic information. To maximize the use of semantic information of training images, we first propose a patch-level graph neural network and transform-based complementary learning module to augment the complementary learning between the various patches in the same class from different images.

\item We leverage a soft-complementary loss function matched with the network. We use the output-patches-combined attention maps generated in the previous training epoch as the reference maps to complement the new generated attention maps and make the newly generated attention maps smoother and more precise.

\item The experiments on the PASCAL VOC 2012 semantic segmentation benchmark demonstrate that we produce more precise pseudo semantic segmentation labels. Besides, we also have good results in the fully supervised semantic segmentation task using the pseudo labels we produce. 
\end{itemize}

\begin{figure*}[ht]
\centering
\includegraphics[width=0.85\linewidth,height=6.5 cm]{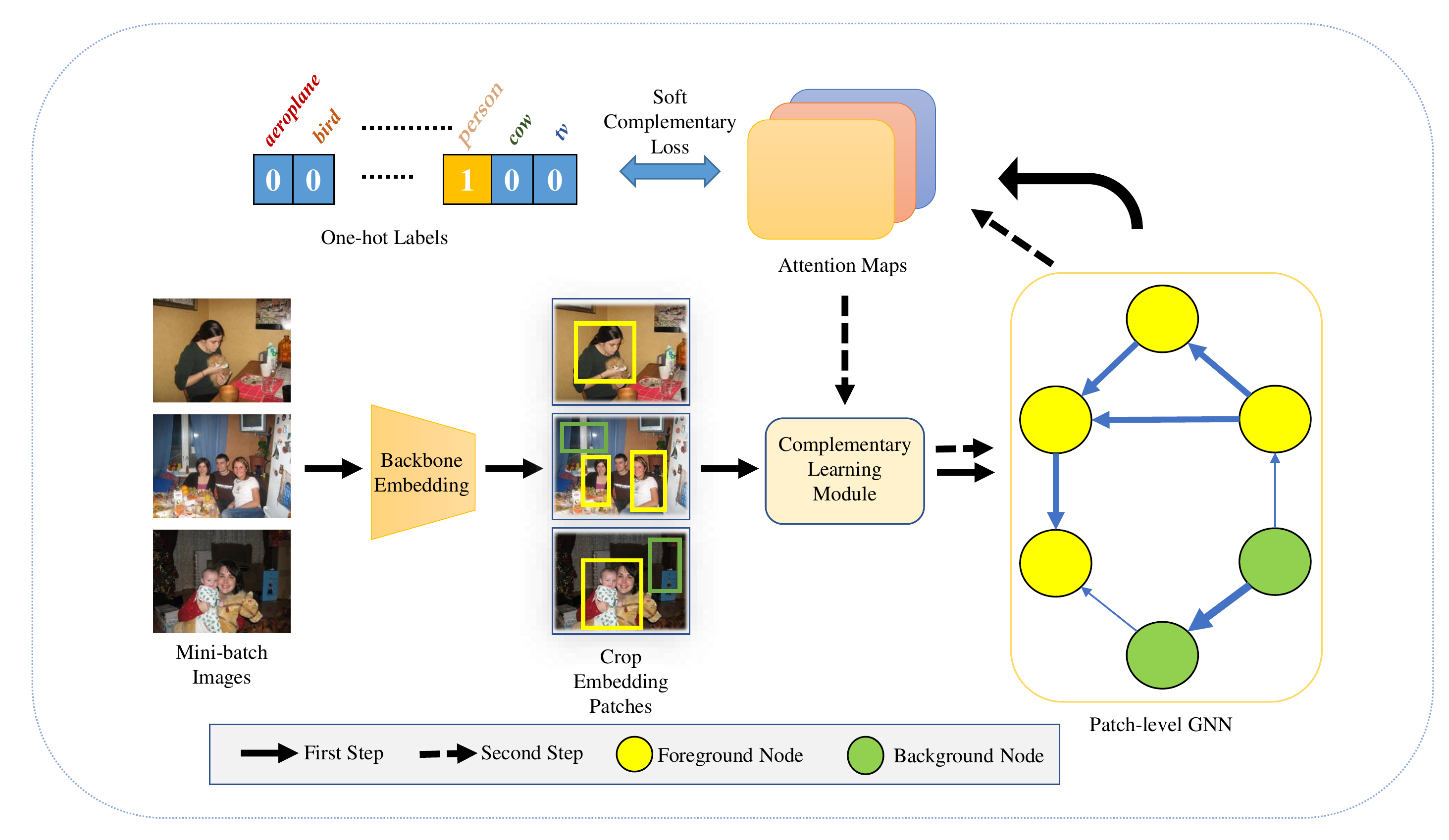} 
\caption{The pipeline of our P-GNN network. The input images are the mini-batch images from the same class. Backbone embedding part is the traditional VGG-16 classification network. Complementary learning module is used to construct P-GNN network. The second step uses attention maps generated in the first step to form the soft complementary loss functions.}
\end{figure*}

\section{Related Work}
\subsection{Weakly supervised semantic segmentation with image-level supervision}
Weakly supervised semantic segmentation (WSSS) with only image-level labels has drawn a lot of attention since this technique greatly frees up the cost of manpower labeling. Mainstream methods in WSSS can be classified into two parts: supplement of semantic information and refinement of edge features. The first type of method tries to fill in missing semantic information in CAMs \cite{zhou2016learning} since CAMs are weak to locate the complete semantic areas. \cite{zhang2018adversarial} erase the most discriminative parts in one image and explore other semantic areas. Other efforts address this difficulty by using dilated convolution \cite{wei2018revisiting} and generating seed regions \cite{huang2018weakly} in one single image. \cite{fan2020cian,sun2020mining} explore semantic relations between pairs of images. The second type of method refines the boundary areas by generating boundary network \cite{ahn2019weakly} and boundary labels \cite{chen2020weakly}.\\
In this paper, we mainly follow the first type and supplement semantic information of multi-patches and multi-images in the same categories. Our model maximally makes up for the lack of pixel-level semantic information in weakly supervised semantic segmentation. Compared with previous methods that concentrate on a single image and pair-wise images, our method extends the breadth of exploration to different patches of multiple images.

\subsection{Explorations on finding complete semantics in weakly supervised semantic segmentation}
In weakly supervised semantic segmentation, the segmentation maps are incomplete in most cases as only finding the most distinctive areas. Some recent works use auxiliary tasks to provide additional pixel-level supervision that guides the generation of pseudo segmentation labels. \cite{zeng2019joint} firstly adopt joint learning methods and use saliency detection maps from other datasets as supervision. \cite{lee2021railroad,xu2021leveraging} use saliency maps generated from other models as additional labels to supplement semantic areas from CAMs. However, these methods use additional labels, so they can't liberate manpower to the maximum. 
Graph neural network is another effective tool to complement semantic information. However, the way on how to construct nodes and edges need to be further explored. Some methods use graph neural networks to build connections between different pixels \cite{zhang2021affinity,pan2021weakly}, but the models only use the semantic information in only a single image so the semantic segmentation regions are not complete. \\
Our model takes advantage of the connections between patches in different images in the same categories and is based on a transformer-like complementary learning module to create weighted edges. We have expanded the breadth to patch-level congeneric semantic information and introduce a transformer-like structure that has been proven effective to enhance mutual semantic learning.
 Therefore, our model can generate a more complete and robust graph neural network system.

\section{Proposed Approach}
The common pipeline for weakly supervised semantic segmentation with only image-level labels contains two major parts: pseudo label generation and semantic segmentation relying on pseudo-ground-truth labels. Our model also includes these two sections. To improve the accuracy of segmentation results comprehensively, there are three aspects we need to consider: (1) How to make full use of semantic information of the same categories in training images to make up for the lack of pixel-level ground-truth information. (2) How to reduce the mislabeled pixels in generated semantic attention maps. (3) How to generate more complete semantic predictions of semantic segmentation network in the second part with different pseudo-ground-truth labels. In this section, we will elaborate our model to solve the corresponding problems of the three aspects above. \\ 
Our model contains two main sections:
(1)	To make up for the lack of spatial information in weakly supervised learning, we generate patch-wise graph neural network and transformer-based complementary learning module to explore the congeneric semantic information among patches with the same class labels in different images. Besides, to refine the semantic maps, we propose one new soft-complementary loss function to reduce mislabeled pixels. This training section can maximize the use of semantic information in training datasets.
(2)	We train fully supervised semantic segmentation using pseudo semantic segmentation labels in the first section and we propose a mutual complementary module to generate more complete semantic predictions. 
This module can be regarded as an auxiliary module of the entire model. 

\subsection{Overview}
To complement each other with semantic information among images in the same class, we generate a patch-level graph neural network in Figure 2 with annotations of class labels. For each mini-batch $\{{I_i}\}_{i=1}^{N}$, we set the input images from the same class. To maximally exploit the congeneric semantic information, we crop each image into different patches and try to distinguish the background patches and foreground patches. The embedding features of each patch construct the node features and the embedding features from other patches serve as “sibling nodes” to provide more reference semantic information of the same category. We feed the patches into our complementary learning module to establish directed edges between patch nodes. Edges represent the similarity between different nodes and the weights on the edges become higher if the model judge that the features of the connected two nodes are similar. Through this module, we construct the P-GNN model and finally fuse the nodes’ features to generate attention maps. With the attention maps generated in the previous epoch as references, we provide the soft-complementary loss function to guide the generation of new attention maps.

\subsection{Construct the patch-level graph neural network}
\subsubsection{The method of cropping images}
The basic method of cropping images into patches is to divide the images evenly. For one embedding feature $\{{F_i}\}$ (w*h*3), we crop it evenly to S*S patches. Patches are all square and have no-overlapping areas. We treat the embedding vector of each patch $\{{P_j}\}$ as the node $\{{N_j}\}$ in the graph. 
\begin{eqnarray}
N_j = F_E(P_j)
\end{eqnarray}
In Eq. 1, $F_E$ is the embedding function, which is comprised of convolutional neural networks.
Although each patch may not contain the entire objects, this simple method makes the foreground and background pixels have a relative distinction and improve the quality of pseudo training labels to some extent. \\
Another cropping method is like the object detection methods. Each patch is the detected region. Here we adopt the weakly supervised object detection method, which only uses class labels like ours. We choose the top k score proposals as node vectors and these proposals have overlapping areas. We can suppose these proposals detect foreground objects, and we treat other undetected areas as one background node $\{{N_b}\}$. 

\begin{figure}[t]
\centering
\includegraphics[width=0.85\linewidth,height=7.5cm]{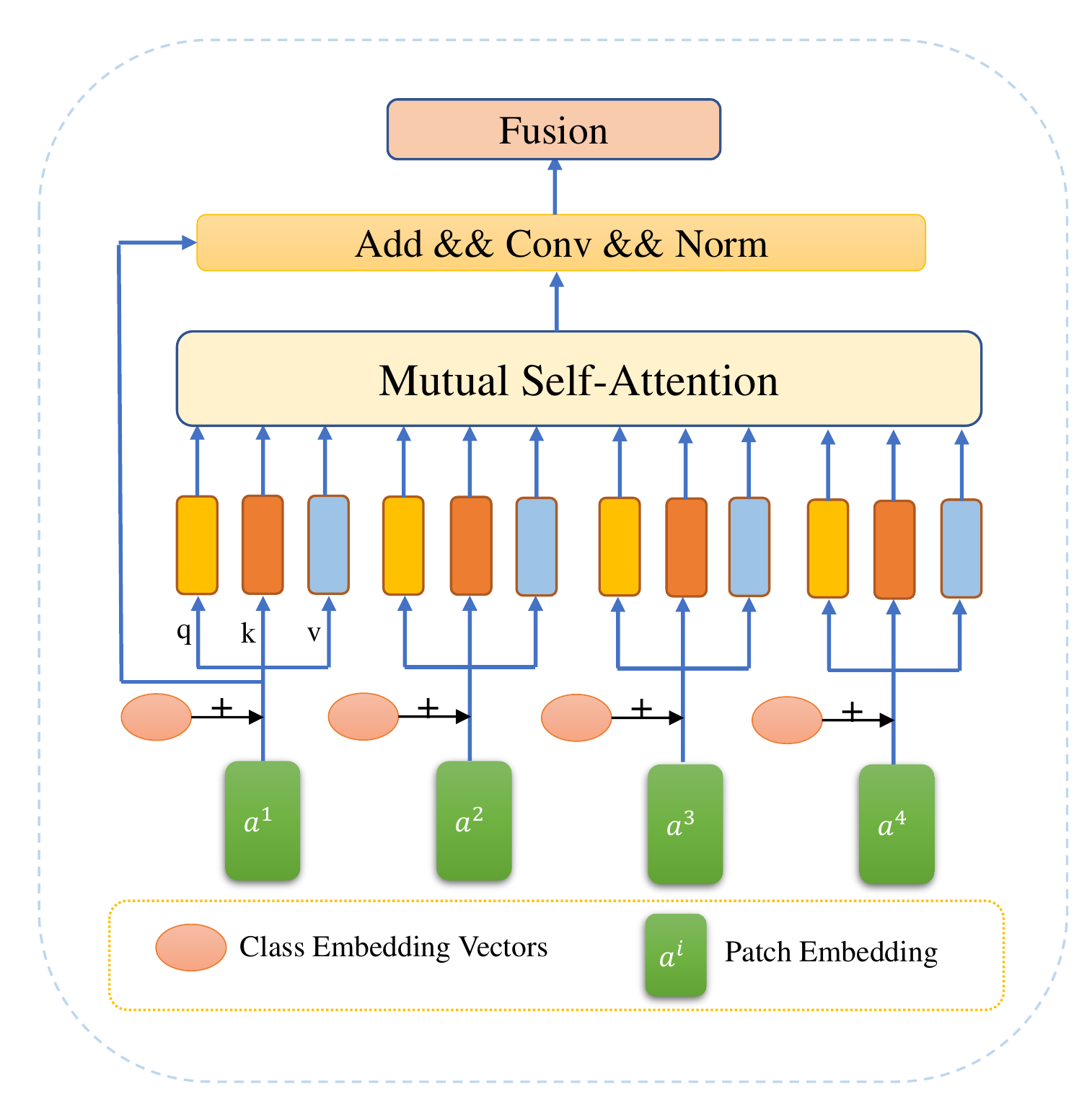} 
\caption{Diagram of the complementary learning module. Compared with the traditional transformer module, we split patches as a sequence and add class embedding vectors.}
\label{trans}
\end{figure}

\subsubsection{Complementary learning module}
To generate edges, we design this transformer-like module to broadly establish weighted connections between different nodes. For each node, we generate its embedding node vector $\{{N_j}\}$ and its corresponding vectors: queries, keys, and values. Like the traditional transformer module \cite{vaswani2017attention}, we calculate the mutual self-attention mechanism for the matrix Q,K,V in Eq. \ref{attention}. 
\begin{eqnarray}
Attention(Q,K,V) = softmax(QK^T/\sqrt{(d_k)})V
\label{attention}
\end{eqnarray}
Queries and keys of dimension $d_k$, and values of dimension $d_v$. After the mutual self-attention module, we add the residual module, convolutional network, and layer normalization to generate a complete transformer-based module shown in Figure \ref{trans}. We can treat the transformer-based module as a weighted matrix W and we can generate weighted edges $E_{\alpha,\beta}$ by Eq. \ref{edge}, which is a simple expression.
\begin{eqnarray}
E_{\alpha,\beta} = {N_\alpha}W{N_\beta}, &\alpha,\beta \in j
\label{edge}
\end{eqnarray}
However, compared to the transformer calculated in NLP, our module calculates the sequence correlation according to the order in which the patches are fed to the model. Therefore, our complementary learning module can establish the connections between the current patch and patches in this image or in other images from the same mini-batch. \\
Besides, we innovatively generate the class-embedding module to substitute the positional embedding module in traditional transformer. For each detected patch, we assign it to the class labels known. For background nodes, we assign their class labels to 0 in Eq. \ref{class}. 
\begin{eqnarray}
Class(N_j)=\left\{
\begin{aligned}
Label(N_j) &,& Foreground \\0&,& Background\\
\end{aligned}
\right.
\label{class}
\end{eqnarray}
We add the class embedding results to the first embedding results after the basic classification network shown in Figure \ref{trans}. This class embedding module serves to enhance the spatial semantic information of pixels, which is a unique design for weakly supervised learning as the lack of spatial information. \\
To meet the requirements of semantic segmentation in computer vision, the patches originally belonging to one image are spliced together to form the final segmented attention maps. 

\begin{figure*}[t]
\centering
\includegraphics[width=0.95\linewidth,height=6.5 cm]{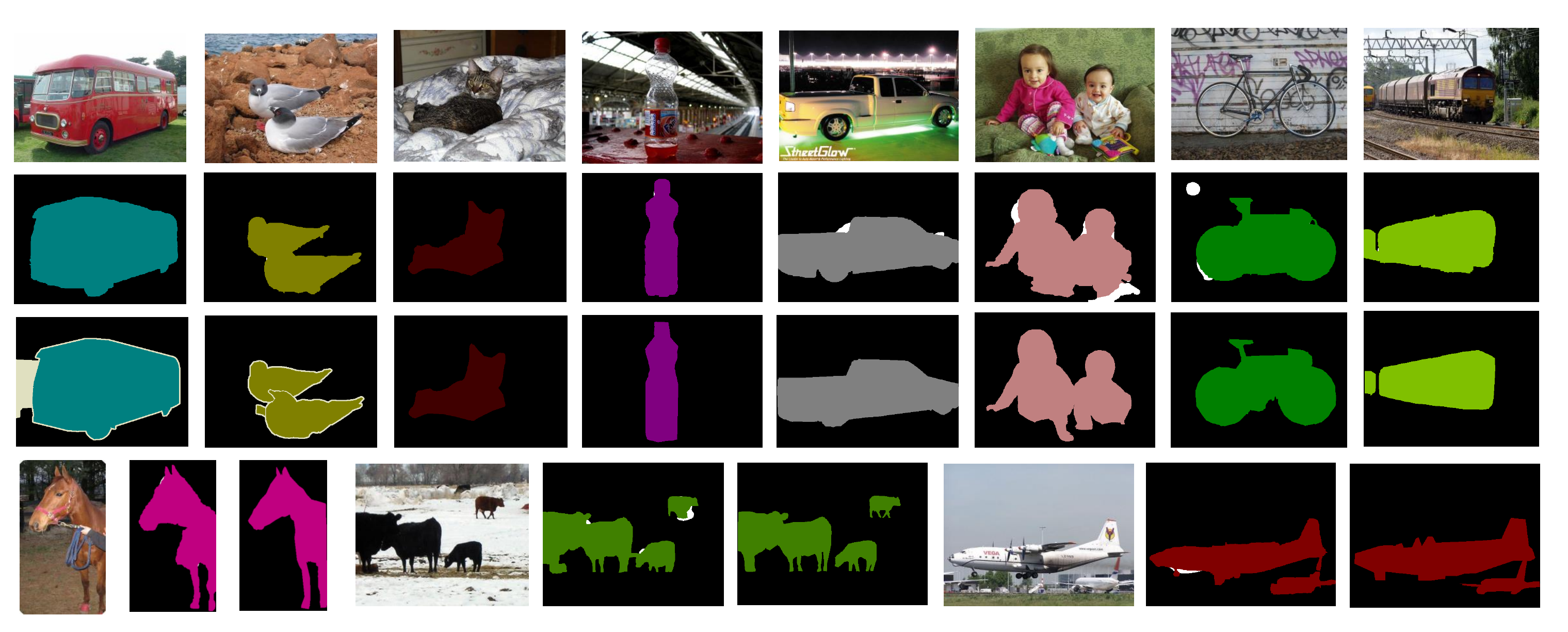} 
\caption{Semantic segmentation results from different classes. Each group contains one image and two segmentation maps. The first segmentation map is generated by our model and the second one is the ground-truth map.}
\label{segmentaion}
\end{figure*}

\subsection{Soft Complementary Loss}
After the patch-level graph neural network, we get the attention maps for each node. We set a threshold for attention maps $\{{A_k}\}$ and get attention regions in Eq. \ref{region}. 
\begin{eqnarray}
Thre(A_k)=\left\{
\begin{aligned}
A_k &,& if (A_k > Threshold) \\ 0&,& else\\
\end{aligned}
\right.
\label{region}
\end{eqnarray}
k represents different classes.\\
With the attention regions, we generate a soft complementary loss function to complement attention maps with two steps' training. The first step of training is what we have presented. After the generation of first attention maps, we adopt the adversarial erasing method, which means we erase the first attention areas in original maps and reenter the maps into our patch-level graph neural network. In Eq. \ref{loss}, we use four parts of losses to make the attention maps detect correct semantic areas. In experiments, we use attention maps generated at the previous epoch as the first step attention maps.
\begin{eqnarray}
\label{loss}
\nonumber L(A)=\lambda_1 l_D(1-A,N)-\lambda_2 l_P(A,N)+\\
\lambda_3 l_{TV}(A)+ \lambda_4 l_c
\end{eqnarray}
Our attention maps are established for every single category. The loss function is designed to make the attention maps of each category as semantically correct as possible. This purpose is consistent with our main idea that makes maximal exploration on correct semantic regions. 
\begin{eqnarray}
l_D(1-M,X)=logP(y=c^*|\phi(1-A,N))
\label{loss1}
\end{eqnarray}
\begin{eqnarray}
l_P(M,X)=logP(y=c^*|\phi(M,N))
\label{loss2}
\end{eqnarray}
In Eq. \ref{loss1} and \ref{loss2}, we erase the attention regions in the original nodes' features and try to explore other areas that also contain the same class semantic information. The second term is designed to keep the attention areas detected in the second step relatively small. 
\begin{eqnarray}
l_{TV}(A)=\sqrt{\sum_{x,y}(A_{x,y}-A_{x,y+1})^2+\sum_{x,y}(A_{x,y}-A_{x+1,y})^2}
\label{loss3}
\end{eqnarray}
The third term is designed to make the attention maps smoother. The first two terms form the complementary loss function. The third term of loss function makes the complementary effects "soft", which means the difference between the values of adjacent pixels is controlled.
\begin{eqnarray}
l_{c}=l_{CE}(P(y=c^*),l_{gt}(c))
\label{loss4}
\end{eqnarray}
The fourth term is used to generate cross-entropy loss of predicted class labels and ground-truth class labels. $l_{c}$ means the classification loss. $l_{CE}$ means cross-entropy loss and $l_{gt}(c)$ is the ground truth class label for class c.\\
Four terms of loss functions are designed to complement attention maps to be more complete and precise. As we set different constraints to the generated attention maps, we call our loss function "soft-complementary" loss function.

\subsection{Fully supervised semantic segmentation}
\begin{figure}[t]
\centering
\includegraphics[width=0.9\linewidth,height=5cm]{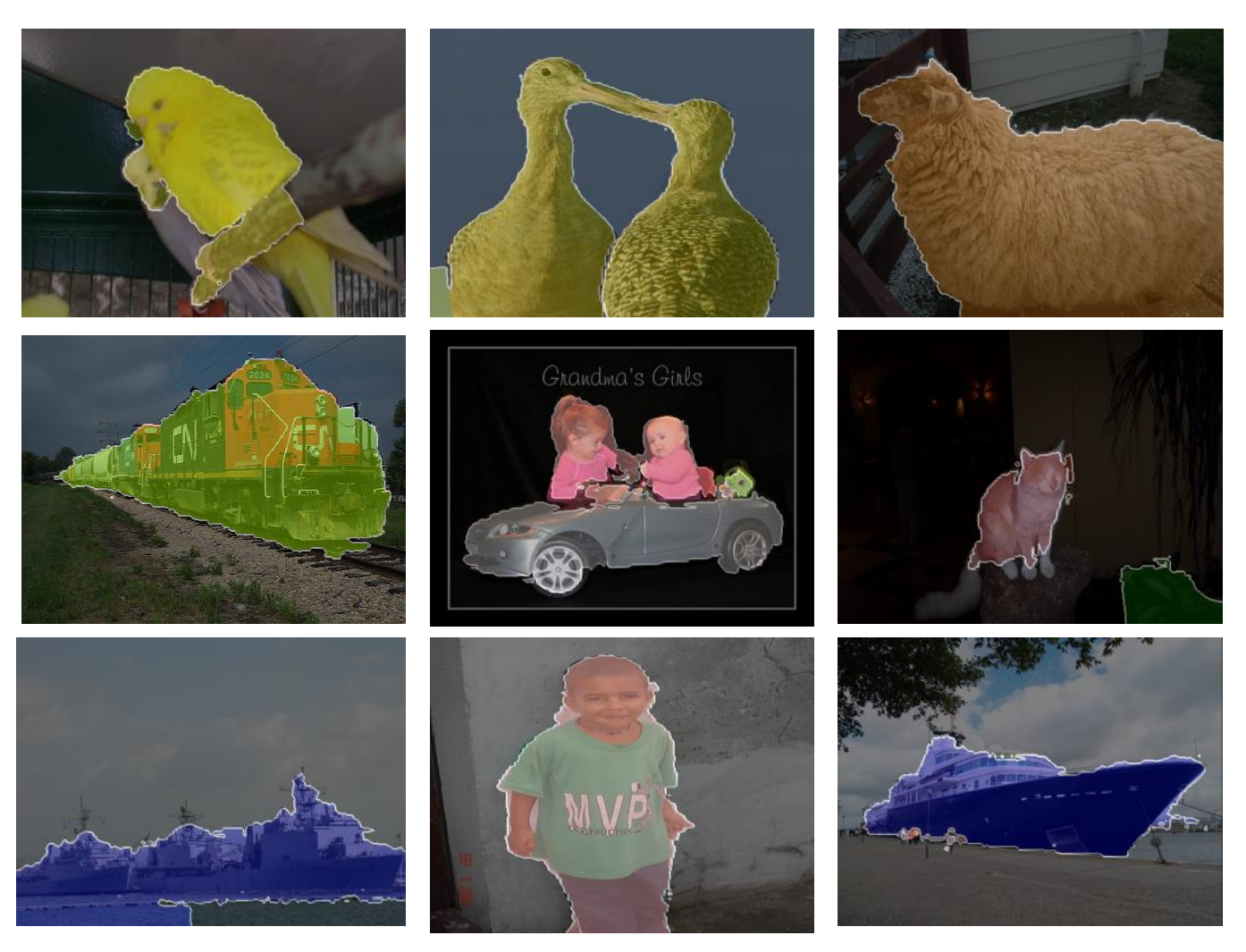} 
\caption{Visualization of some pseudo labels generated by fully supervised semantic segmentation network.}
\label{sub}
\end{figure}

After the generation of pseudo semantic segmentation labels, we use Deeplab-v2 \cite{chen2017deeplab} to train the fully supervised semantic segmentation network. \\
To improve the quality of pseudo-ground-truth labels, we propose a mutual complementary module to update the pseudo labels at each epoch. In Eq. \ref{mutual}, $C_{x,y,t}$ is the pseudo-class label of one pixel at location (x,y) at the epoch t. $\phi_{x,y,t+1}$ is the prediction label by convolutional neural network of this pixel at location (x,y) at the epoch t+1. '0' means background. The core idea is to fill in the predicted foreground pixels as much as possible. After observing the predicted labels, we find the main problem that limits the accuracy of pseudo labels is the incorrect background pixels, which means the model always misclassifies foreground pixels as background pixels. Due to this limitation, we refine our pseudo labels with reference to previous labels to complement semantic information. \\
Some visualization results of pseudo labels generated by Deeplab-v2 with our mutual complementary module are shown in Figure \ref{sub}.

\begin{eqnarray}
C_{x,y,t+1}=\left\{
\begin{aligned}
C_{x,y,t} &,& if (\phi_{x,y,t+1} =0) \\ \phi_{x,y,t+1} &,& if (C_{x,y,t} =0) \\ \phi_{x,y,t+1} &,& else
\end{aligned}
\right.
\label{mutual}
\end{eqnarray}

\section{Experiment}
\subsection{Experimental Details}
\subsubsection{The Dataset in Semantic Segmentation}
We train and test our model on PASCAL VOC 2012 dataset. With image-level category labels for training, we test the quality of pseudo labels we produce on the PASCAL VOC train set by comparing mIoU with ground truth labels. This dataset has 1464 images for training, 1449 images for validation, and 1456 images for testing. Moreover, we leverage data augmentation results in the previous work \cite{AUG} and have 10582 images for training. 

\subsubsection{Performance Evaluation Measures}
To conduct a quantitative performance evaluation, we use mIoU to evaluate the quality of produced pseudo semantic segmentation labels and test semantic segmentation results. The calculation function is shown in Eq. \ref{miou}.
\begin{eqnarray}
 mIoU = \frac{TP}{TP+FP+FN} 
 \label{miou}
\end{eqnarray}
TP, FP, FN mean true-positive labels, false-positive labels, and false-negative labels, respectively. \\
At the same time, as our main contribution is focused on reducing false-negative samples, we further compare the  $precision$ and $recall$ in Eq. \ref{pre} and Eq. \ref{rec}.
\begin{eqnarray}
precision = \frac{TP}{TP+FP} 
\label{pre}
\end{eqnarray}
\begin{eqnarray}
recall = \frac{TP}{TP+FN} 
\label{rec}
\end{eqnarray}

\subsubsection{Training Details}
As for the hyperparameters $\lambda_i$ in our loss function, we set them to different values to balance the weights of different terms. We don't try all the possible values of $\lambda_i$. We set them to 1,1,0.8,1 separately to keep a relative balance. For each mini-batch, we set the input images to 3. 

\subsubsection{Backbone Settings}
We adopt VGG-16 and ResNet-101 as the backbone in our patch-level graph neural network and compare the performance with the methods using the same backbone as ours.

\subsection{Precision and Recall comparison}
To show the contribution of this paper more prominently, we compare our semantic segmentation results with CAMs in $precision$ and $recall$ on PASCAL VOC 2012 train set. 
As our main contribution is reducing false-negative samples, we mainly focus on improving the $recall$ part. 
\begin{table}[t]
\centering
\setlength{\tabcolsep}{5mm}{
\begin{tabular}{ c  c  c }
\hline
        Methods &  mAP & recall \\
\hline
         
        CAMs & 46.3  & 59.7  \\
        Ours & \textbf{57.8} & \textbf{86.3} \\

\hline
\end{tabular}}
\caption{Comparison in mAP and recall of different methods on PASCAL VOC 2012 train set.}
\label{recall}
\end{table}

\begin{table}[t]
\centering
\setlength{\tabcolsep}{0.9mm}{
\begin{tabular}{  c  c  c  c }
\hline
     Method & Sup. &  val & test  \\
\hline

     STC\cite{wei2016stc} & I. & 49.8 & 51.2 \\ 

     SEC\cite{kolesnikov2016seed} & I. & 50.7 & 51.7 \\ 
     
     AugFeed\cite{Qi2016aug} & I. & 54.3 & 55.5 \\
     
     AE-PSL\cite{wei2017a} & I. & 55.0 & 55.7 \\
     
     ESOS\cite{oh2017exploiting} & I. & 55.7 & 56.7 \\
     
     MCOF\cite{wang2018weakly}  & I. & 56.2 & 57.6 \\
     
     AFFNet\cite{ahn2018learning} & I. & 58.4 & 60.5 \\
     
     DSRG\cite{huang2018weakly}   & I. & 59.0 & 60.4 \\
     OAA\cite{jiang2019integral}  & I. & 63.1 & 62.8  \\
     CIAN\cite{fan2020cian}  & I. & 62.4 & 63.8  \\
     
     ICD\cite{fan2020learning}  & I. & 64.0 & 63.9  \\
     GSM\cite{li2020group}  & I. & 63.3 &  63.6  \\
     NSR\cite{yao2021non}  & I. & 65.5 &  65.3  \\
                               
\hline     
     BoxSup\cite{dai2015boxsup} & B. & 62.0 & 64.6 \\
     
     ScribbleSup\cite{lin2016scribblesup} & S. & 63.1 & -  \\
\hline     
     FCN\cite{long2015fully} & F. & - &  62.2 \\   

     DeepLab\cite{chen2017deeplab} & F. & 67.6 & 70.3\\ 
\hline
     Ours & I. & 66.3  & 66.7 \\ 
     Ours + MC & I. & \textbf{66.5}  & \textbf{67.0} \\ 
\hline
\end{tabular}}
\caption{Semantic Segmentation results on VGG-16 backbone on PASCAL VOC 2012 validation and test set. The supervision types (Sup.) indicate: image-level label(I.), bounding box(B.), and fully supervised label(F.). MC is the mutual complementary module we use in fully supervised semantic segmentation.}
\label{seg_VGG}
\end{table}

\begin{table}[t]
\centering
\setlength{\tabcolsep}{0.9mm}{
\begin{tabular}{  c  c  c  c }
\hline
     Method & Sup. &  val & test  \\
\hline

     DCSP\cite{chaudhry2017discovering} & I. & 60.8 & 61.9 \\ 

     DSRG\cite{huang2018weakly} & I. & 61.4 & 63.2 \\

     MCOF\cite{wang2018weakly}  & I. & 60.3 & 61.2 \\
     
     AFFNet\cite{ahn2018learning} & I. & 61.7 & 63.7 \\
     IRN\cite{ahn2019weakly} & I. & 63.5 & 64.8 \\
     FickleNet\cite{lee2019ficklenet}   & I. & 64.9 & 65.3 \\
     OAA\cite{jiang2019integral}  & I. & 65.2 & 66.4  \\
     SEAM\cite{wang2020self}  & I. & 64.5 & 65.7  \\
                    
     CIAN\cite{fan2020cian}  & I. & 64.3  & 65.3  \\
     ICD\cite{fan2020learning}  & I. & 67.8 & 68.0  \\

     GSM\cite{li2020group}  & I. & 68.2 & 68.5  \\
     
     WSGCN-I\cite{pan2021weakly}   & I. & 66.7 & 68.8  \\
     
     NSR\cite{yao2021non}  & I. & 68.3 & 68.5  \\
     
\hline
     Ours & I. & 70.5  & 70.6 \\ 
     Ours + MC & I. & \textbf{70.7}  & \textbf{70.8} \\ 
\hline
\end{tabular}}
\caption{Semantic Segmentation results on ResNet-101 backbone on PASCAL VOC 2012 validation and test set. MC is the mutual complementary module.}
\label{seg_Res}
\end{table}
As shown in Table \ref{recall}, our model shows the superiority in reducing false-negative samples greatly and thus has a higher $recall$ value. Meanwhile, the mAP value of our model is still higher than that of CAMs by 11.5$\%$. Our patch-level graph neural network can mainly reduce false-negative mislabeled pixels by augmenting mutual learning of different patches in the same class. Through experimental results, we find our model can improve mAP value to some extent as the augmentation can refine object boundaries relatively. \\
Through the comparison in Table \ref{recall}, we show the superiority of the model itself compared with traditional CAMs. In the ablation study part, we will analyze the effects of different modules separately.

\subsection{Comparisons with State-of-the-Arts in Semantic Segmentation}
We train DeepLab v2-ResNet101 \cite{chen2017deeplab} with our pseudo semantic segmentation labels to test our model on the semantic segmentation task. The semantic results are still great and can surpass previous state-of-the-art methods on weakly supervised semantic segmentation. We have quantitative results on PASCAL VOC 2012 validation and test set in Tables \ref{seg_VGG} and \ref{seg_Res}. Besides, we show different comparison results under different backbones. Based on the VGG-16 backbone, we compare our results with some methods using the same image-level labels as us on PASCAL VOC 2012 validation set and test set. Our methods are even comparable with some methods using stronger supervision labels, like BoxSup (trained with bounding boxes), and ScribbleSup (trained with scribbles). Compared to some advanced methods published in 2021, like GSM \cite{li2020group} and NSR \cite{yao2021non}, our model has better performance using the same backbone. We lead GSM by 3.0$\%$ and 3.1$\%$ separately and lead NSR by 1.0$\%$ and 1.7$\%$ separately on PASCAL VOC 2012 validation set and test set. 
Figure \ref{segmentaion} visualizes some semantic segmentation results obtained by our approach. We choose semantic segmentation results from different classes to show the good and robust performance of our model. \\
Based on the ResNet-101 backbone, we compare methods using the same image-level labels in Table \ref{seg_Res}. Our model surpasses NSR by 2.4$\%$ and 2.3$\%$ on the validation set and test set. In all, our model shows good performance under different backbones.

\begin{table}[t]
\centering
\setlength{\tabcolsep}{5mm}{
\begin{tabular}{ c  c  c }
\hline
         &  mAP & recall \\
\hline
         
        CAMs & 46.3  & 59.7  \\
        CAMs + P-GNN + $l_c$ loss & 57.3 & 85.6 \\
        CAMs + P-GNN + SC loss & \textbf{57.8} & \textbf{86.3} \\

\hline
\end{tabular}}
\caption{Results in mAP and recall in our model on PASCAL VOC 2012 train set. P-GNN is short for our patch-level graph neural network. $l_c$ loss is the classification cross-entropy loss. SC loss is the soft-complementary loss in our model.}
\label{ablation}
\end{table}

\begin{table}[t]
\centering
\setlength{\tabcolsep}{5mm}{
\begin{tabular}{ c  c  c }
\hline
        Number of Patches &  mAP & recall \\
\hline
         
       4 & 57.1  & 82.3  \\
       
       16 & \textbf{57.8} & \textbf{86.3} \\
       36 & 56.8 & 81.5 \\
       64 & 55.4 & 78.2 \\

\hline
\end{tabular}}
\caption{Experiments on different patch (node) numbers on PASCAL VOC 2012 train set.}
\label{patch}
\end{table}

\subsection{Ablation Study}
In this part, we analyze the independent effects of each module in reducing false negative pixels. As $precision$ and $recall$ should be balanced in comparison, we show both of them in Table \ref{ablation}. We can find our patch-level graph neural network improves the $recall$ value by 25.9$\%$ and improve the $precision$ value by 11.0$\%$. This improvement reflects the effectiveness of our method numerically. Besides, our soft-complementary loss can improve the $recall$ value by 0.5$\%$ and improve the $precision$ value by 0.7$\%$. Although the corresponding improvement is relatively small, the loss function design shows its effectiveness in improving $recall$.

The number of patches we set in our P-GNN influences the learning effectiveness and complexity of the entire network. The hyperparameter is really important and we analyze different numbers in Table \ref{patch}. We find that too many or too few patches will affect the overall training effects. If we set the patch number to 4 or more than 36, the quality of pseudo labels on the training set declines on both $precision$ and $recall$ aspects. If the number of patches increases excessively, the discrimination of patches decreases, which is not conducive to learning. The most suitable patch number is 16 in our experiments. 

As for different $\lambda$ values in our soft-complementary loss function, we set them for different values and have the experimental results in Table \ref{lambda}. We find the first two terms keep preserving and erasing effects to be balanced. If we change $\lambda_1$ or $\lambda_2$ to lower values, both mAP and $recall$ decrease. However, if we change $\lambda_3$ to lower values, both mAP and $recall$ increase as the third term is a strong constraint on pixels' values. Although there are slight changes in experimental results, this group of experiments give us instructions on a deep understanding of each loss term.

\begin{table}[ht]
\centering
\setlength{\tabcolsep}{4mm}{
\begin{tabular}{ c c c c   c  c }
\hline
         $\lambda_1$  & $\lambda_2$  & $\lambda_3$  & $\lambda_4$  &  mAP & recall \\
\hline
        1 & 1 & 1 & 1  & 57.4 & 86.2 \\
        1 & 1 & 1 & 0.8 & 57.6  & 86.1  \\
        1 & 1 & 0.8 & 1  & \textbf{57.8} & \textbf{86.3} \\
       1 & 0.8 & 1 & 1  & 57.4 & 85.9 \\
      0.8 & 1 & 1 & 1  & 57.3 & 85.8 \\

\hline
\end{tabular}}
\caption{Comparison in mAP and recall of different $\lambda$ values in our soft-complementary loss function on PASCAL VOC 2012 train set.}
\label{lambda}
\end{table}

\section{Conclusion}
In this paper, we are the first to explore the congeneric semantic segmentation regions in the same class through a patch-level graph neural network. In exploring the complementary learning of congeneric images, our innovation lies in exploring from a more detailed patch level and construct effective graph neural network. We take embedding patches as nodes and use a complementary learning module to construct weighted edges. Besides, we design a soft-complementary loss function to further reduce  incorrectly classified pixels. Our P-GNN generates pseudo labels with high $recall$ scores and train Deeplab-v2 with our mutual complementary module to finish semantic segmentation. \\
In all, our method supplements object semantic features from many aspects and has good performance in Pascal VOC 2012 dataset.

\bibliography{aaai22.bib}  
\end{document}